\documentclass[twocolumn]{article}

\usepackage[width=17cm,height=22cm]{geometry}
\usepackage[french,english]{babel}
\usepackage[utf8]{inputenc}
\usepackage{fancyvrb}
\usepackage{authblk}
\usepackage{hyperref}

\usepackage{graphicx}

\title{Disease Classification in Metagenomics with 2D Embeddings and Deep Learning}
\author[1]{Thanh Hai Nguyen \thanks{nthai@cit.ctu.edu.vn}}
\author[1]{Edi Prifti}
\author[3]{Yann Chevaleyre}
\author[4]{Nataliya Sokolovska}
\author[2]{Jean-Daniel Zucker}
\affil[1]{Integromics, Institute of Cardiometabolism and Nutrition. Paris, France}
\affil[2]{Ummisco, IRD, Sorbonne Université, Bondy, France}
\affil[3]{LAMSADE, Université Paris Dauphine, Paris, France}
\affil[4]{Nutriomics, Sorbonne Université, INSERM, Hôpital la Pitié Salpêtrière, Paris, France}

\begin{document}
\maketitle

\begin{abstract}
Deep learning (DL) techniques have shown unprecedented success when applied to images, waveforms, and text. Generally, when the sample size ($N$) is much bigger than the number of features ($d$), DL often outperforms other machine learning (ML) techniques, often through the use of Convolutional Neural Networks (CNNs). However, in many bioinformatics fields (including metagenomics), we encounter the opposite situation where $d$ is significantly greater than $N$. In these situations, applying DL techniques would lead to severe over-fitting. 

Here we aim to improve classification of various diseases with metagenomic data through the use of CNNs. For this we proposed to represent metagenomic data as images. The proposed \textbf{Met2Img} approach relies on taxonomic and t-SNE embeddings to  transform abundance data into "synthetic images". 

We applied our approach to twelve benchmark data sets including more than 1400 metagenomic samples. Our results show significant improvements over the state-of-the-art algorithms (Random Forest (RF), Support Vector Machine (SVM)). We observe that the integration of phylogenetic information alongside abundance data improves classification. The proposed approach is not only important in classification setting but also allows to visualize complex metagenomic data. The Met2Img is implemented in Python. 

\end{abstract}

\section{Introduction}
\label{sec:Introduction}
High throughput data acquisition in the biomedical field has revolutionized research and applications in medicine and biotechnology. Also known as ``omics'' data, they reflect different aspects of systems biology (genomics, transcriptomics, metabolomics, proteomics, etc.) but also  whole biological ecosystems acquired through metagenomics. There is an increasing number of datasets which are publicly available. Different statistical methods have been applied to classify patients from controls\cite{Ginsburg2009-pa} and some have also performed meta-analyses on multiple datasets \cite{Pasolli2016-dk}. However, exploring omics data is challenging, since the number of features $d$ is very large, and the number of observations $N$ is small. Up to now, the most successful techniques applied to omics datasets have been mainly Random Forest (RF), and sparse regression. 

In this paper, we applied DL on six metagenomic datasets, which reflect bacterial species abundance and presence in the gut of diseased patients and healthy controls. We also evaluate our method on additional datasets with \textit{genus} abundance. Since DL performs particularly well for image classification, we focused in this work in the use of CNNs applied to images. For this purpose, we first searched for ways to represent metagenomic data as "synthetic" images and second applied CNN techniques to learn representations of the data that could be used next for classification purposes. 

There are numerous studies where machine learning (ML) is applied to analyze large metagenomic datasets. \cite{Pasolli2016-dk} proposed a unified methodology, MetAML, to compare different state-of-the-art methods (SVM, RF, etc.) on various metagenomic datasets. These datasets were processed using the same bioinformatics pipeline for comparative purposes. A general overview of ML tools for metagenomics is provided by \cite{Soueidan2017-eb}. 
In \cite{Ditzler2015-yo}, a typical pipeline for metagenomic analysis was suggested. It includes data preprocessing, feature extraction, followed by classification or clustering. In order to extract relevant feature, raw data can be arbitrarily transformed. A new shape of data is supposed to help predictors to predict more accurately. In some cases, if domain knowledge is available, it is possible to introduce this prior knowledge into the learning procedure. It can be a heuristic and/or hand picking to select a subset of most relevant features. \cite{Ditzler2015-yo} proposed to use deep belief networks and recursive neural networks for metagenomics. 

A metagenomic dataset usually contains bacterial DNA sequences obtained from an ecosystem such as the intestinal microbiome for instance. These sequences are usually transformed using bioinformatics pipelines into a vector representing the abundance of a set of features (species or other taxonomic groups). One approach of conversion from sequences to vectors removed short and low quality score reads from the whole collection of sequences. Then, the remainder is clustered with CD-HIT \cite{Li2006-sf}, for example into non redundant representative sequences - usually called a catalog. Reads are typically aligned to the catalog and counted. Annotation algorithms are applied to the catalog allowing to infer abundance on different taxonomic levels. 
Other groups have also tackled the use of DL in classifying metagenomic data. For instance, some authors introduced approaches to design a tree like structure for data sets of metagenomics learned by neural networks. A study in \cite{dai_multi-cohort_2018} identified a group of bacteria that is consistently related to Colorectal Cancer (COL) and revealed potential diagnosis of this disease across multiple populations and classified COL cases with a SVM model using the seven COL-enriched bacterial species. The authors in \cite{Fioravanti2017-tr} introduced a novel DL approach, Ph-CNN, for classification of metagenomics using hierarchical structure of Operational Taxonomic Unit (OTU). The performance of Ph-CNN is promising compared to SVM, RF and a fully connected neural network. 

Image classification based on ML has obtained great achievements and there are numerous studies attempting to propose architectures to improve the performance. Alex et al. in \cite{Alex_Krizhevsky2012-ja} introduced a deep and big CNN (including 60 million parameters) which performed on a very large dataset with 1.2 million color images of 224x224 in ImageNet LSVRC-2010. The architectures consisted of 5 convolutional and 3 fully connected layers. The network achieved top-1 and top-5 error rates of 37.5$\%$ and 17$\%$ , respectively.  The authors in \cite{Zeiler2014-ft} presented a novel technique (ZFNet) for visualizing feature maps through convolutional layers, and investigated how to improve the performance of Convolutional Neural Networks (convnet).  A team at Google in 2014 proposed GoogLeNet \cite{GoogLeNet_2014}, a very deep architecture of CNN with 22 layers that won the ImageNet Large-Scale Visual Recognition Challenge 2014 (ILSVRC14). In \cite{Karen_Simonyan2015-mh}, the authors presented how depth affects the performance of CNNs (VGGNet). The authors designed deep architectures of CNNs with very small convolutions (3x3) and achieved top-1 validation error and top-5 validation error were 23.7$\%$  and 6.8$\%$ ,respectively. The authors in \cite{He2015-yb} introduced a residual learning framework (ResNet) that was up to 152 layers but being less complex. ResNet won the 1st places on the ILSVRC 2015 and COCO 2015 competitions. 

Here, we present the \textbf{Met2Img} framework based on \textit{Fill-up} and \textit{t-SNE} to visualize features into images. Our objectives were to propose efficient representations which produce compact and informative images, and to prove DL techniques as efficient tools for prediction tasks in the context of metagenomics. Our contribution is multi-fold:
%
%

\begin{itemize}
	\item We propose a \textit{visualization approach for metagenomic data} which shows significant results on 4 out of 6 datasets (Cirrhosis, Inflammatory Bowel Disease (IBD), Obesity, Type 2 Diabetes) compared to MetAML \cite{Pasolli2016-dk}.\par

	\item CNNs outperform standard shallow learning algorithms such as RF and SVM. This proves that deep learning is a promising ML technique for metagenomics.\par

	\item We illustrate that the proposed method not only performs competitively on species abundance data but also shows significant improvements compared to the state-of-the-art, Ph-CNN, on genus-level taxonomy abundance with six various classification tasks on IBD dataset published in \cite{Fioravanti2017-tr}.
\end{itemize}\par

The paper is organized as follows. In Section \ref{sec:datasets} and \ref{sec:Methodology}, we describe the benchmarks and methods to visualize feature abundance with the Fill-up and t-SNE approaches. We show our results in Section \ref{sec:Results} for different deep learning architectures, for 1D and 2D representations. At the end of this section, we illustrate the performance of our approach on another additional dataset consisting of six classification tasks on IBD with genus-level abundance. Conclusion is presented in the Section \ref{sec:conclusion}.
\section{Metagenomic data benchmarks}
\label{sec:datasets}
\begin{table*}[t]
 			\centering
\begin{tabular}{|l|c|c|c|c|c|c|c|c|c|c|c}
\hline
 \textbf{Group A} & & & & & & \\ \hline

 Dataset name &	\textbf{CIR }&	\textbf{COL} &	\textbf{IBD} &	\textbf{OBE} &	\textbf{T2D} &	\textbf{WT2} \\ \hline 
\#features &	542 &	503 &	443 &	465 &	572 &	381 \\ \hline 
\#samples &	232 &	121 &	110 &	253 &	344 &	96 \\ \hline 
\#patients &	118 &	48 &	25 &	164 &	170 &	53 \\ \hline 
\#controls &	114 &	73 &	85 &	89 &	174 &	43 \\ \hline 
ratio of patients&	0.51 &	0.40 &	0.23 &	0.65 &	0.49 &	0.55 \\ \hline 
ratio of controls &	0.49 &	0.60 &	0.77 &	0.35 &	0.51 &	0.45 \\ \hline 
Autofit size for images &	24$\times$24 &	23$\times$23 &	22$\times$22 &	22$\times$22 &	24$\times$24 &	20$\times$20 \\ \hline 
\hline
 \textbf{Group B} & & & & & & \\ \hline
  Dataset name &	\textbf{CDf} &	\textbf{CDr} &	\textbf{iCDf} &	\textbf{iCDr} &	\textbf{UCf} &	\textbf{UCr} \\ \hline 
\#features &	259 &	237 &	247 &	257 &	250 &	237 \\ \hline 
\#samples &	98 &	115 &	82 &	97 &	79 &	82 \\ \hline 
\#patients &	60 &	77 &	44 &	59 &	41 &	44 \\ \hline 
\#controls &	38 &	38 &	38 &	38 &	38 &	38 \\ \hline 
ratio of patients&	0.61 &	0.67 &	0.54 &	0.61 &	0.52 &	0.54 \\ \hline 
ratio of controls &	0.39 &	0.33 &	0.46 &	0.39 &	0.48 &	0.46 \\ \hline 
Autofit size for images &	17$\times$17 &	16$\times$16 &	16$\times$16 &	17$\times$17 &	16$\times$16 &	16$\times$16 \\ \hline  \end{tabular}
  \caption{Information on datasets}
  \label{tab:inf_data}
\end{table*}
We work with metagenomic abundance data, i.e. data indicating how present (or absent) is an OTU (Operational taxonomic unit) in human gut.
We evaluated our method with each of the different visual representations on twelve different datasets (see Table \ref{tab:inf_data}) with two groups (A and B). Group A consists of datasets including bacterial species related to various diseases, namely: liver cirrhosis (CIR), colorectal cancer (COL), obesity (OBE), inflammatory bowel disease (IBD) and Type 2 Diabetes (T2D) \cite{Pasolli2016-dk}, \cite{Qin2014-hh}, \cite{Zeller2014-ls}, \cite{Le_Chatelier2013-yw}, \cite{Qin2010-pd}, \cite{Qin2012-am}, \cite{Karlsson2013-dr}], with CIR (n=232 samples with 118 patients), COL (n=48 patients and n=73 healthy individuals), OBE (n=89 non-obese and n=164 obese individuals), IBD (n=110 samples of which 25 were affected by the disease) and T2D (n=344 individuals of which n=170 are T2D patients). In addition, one dataset, namely WT2, includes 96 European women with n=53 T2D patients and n=43 healthy individuals. The abundance datasets are transformed to obtain another representation based on feature presence when the abundance is greater than zero. These data were obtained using the default parameters of MetaPhlAn2 \cite{Truong_undated-kb} as detailed in Pasolli et al. \cite{Pasolli2016-dk}. Group B includes Sokol’s lab  data \cite{Sokol2016-sw} consisting of microbiome information of 38 healthy subjects (HS) and 222 IBD patients. The bacterial abundance includes 306 OTUs with genus level abundance. Patients in this data are classified into two categories according to the disease phenotype Ulcerative colitis (UC) and Crohn’s disease (CD). Each category is divided into two conditions (flare (f), if patients got worse or reappeared the symptoms and remission (r), if patients’ symptoms are decreased or disappear). The dataset was partitioned into subset with ileal Crohn’s disease (iCD) and colon Crohn’s disease (cCD). The detail description of the data was presented in \cite{Fioravanti2017-tr}. 

For each sample, species/genus abundance is a relative proportion and is represented as a real number - the total abundance of all species/genus sums to 1. 
\section{Methodology}
\label{sec:Methodology}
Our approach consists of the following steps: First, a set of colors is chosen and applied to different binning approaches (see \ref{method:bins}). The binning can be performed on a logarithmic scale or a transformation. Then, the features are visualized into images by one of two different ways (phylogenetic-sorting (using \textbf{Fill-up}) or visualized based on t-Distributed Stochastic Neighbor Embedding (t-SNE) \cite{Maaten2008-zb} (see \ref{method:Image_generation}).\  t-SNE technique is  useful to find faithful representations for high-dimensional points visualized in a more compact space, typically the 2D plane. For phylogenetic-sorting, the features which are bacterial species are arranged based on their taxonomic annotation ordered alphabetically by concatenating the strings of their taxonomy (i.e. phylum, class, order, family, genus and species). This ordering of the variables embeds into the image external biological knowledge, which reflects the evolutionary relationship between these  species. Each visualization method will be used to either represent abundance or presence data. The last representation, which serves as control is the 1D of the raw data (with the species also sorted phylogenetically). For the t-SNE representation, we use only training sets to generate global t-SNE maps, images of training and test set are created from these global maps.  

\subsection{Abundance Bins for metagenomic synthetic images}
\label{method:bins}
In order to discretize abundances as colors in the images with, we use different methods of binning. Each bin is illustrated by a distinct color extracted from color strip of heatmap colormaps in Python library such as \textit{jet}, \textit{rainbow}, \textit{viridis}. In \cite{liu_somewhere_2018}, authors stated that \textit{viridis} showed a good performance in terms of time and error. The binning method we used in the project is Unsupervised Binning which does not use the target (class) information. In this part, we use \textbf{EQual Width binning} (EQW) with ranging \textbf{[Min, Max]}. We test with $k = 10$ bins (for color distinct images, and gray images), width of intervals is $w = 0.1$, if Min=0 and Max = 1, for example.

\subsubsection{Binning based on abundance distribution}
\begin{figure} 
	\centering
   \includegraphics[width=0.5\textwidth]{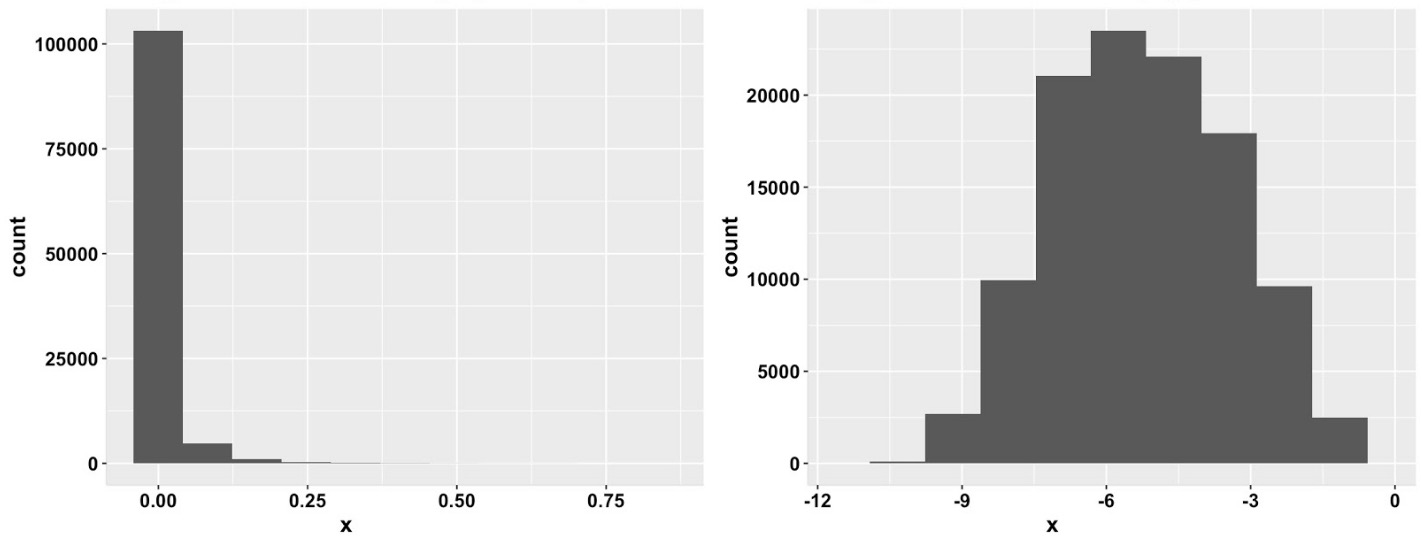}
		\caption{Histogram of whole six datasets of group A (Left: original data, Right: Log-histogram (base 4)}
		\label{fig:hiswhole6}
\end{figure}
\begin{figure} 
	\centering
   \includegraphics[width=0.5\textwidth]{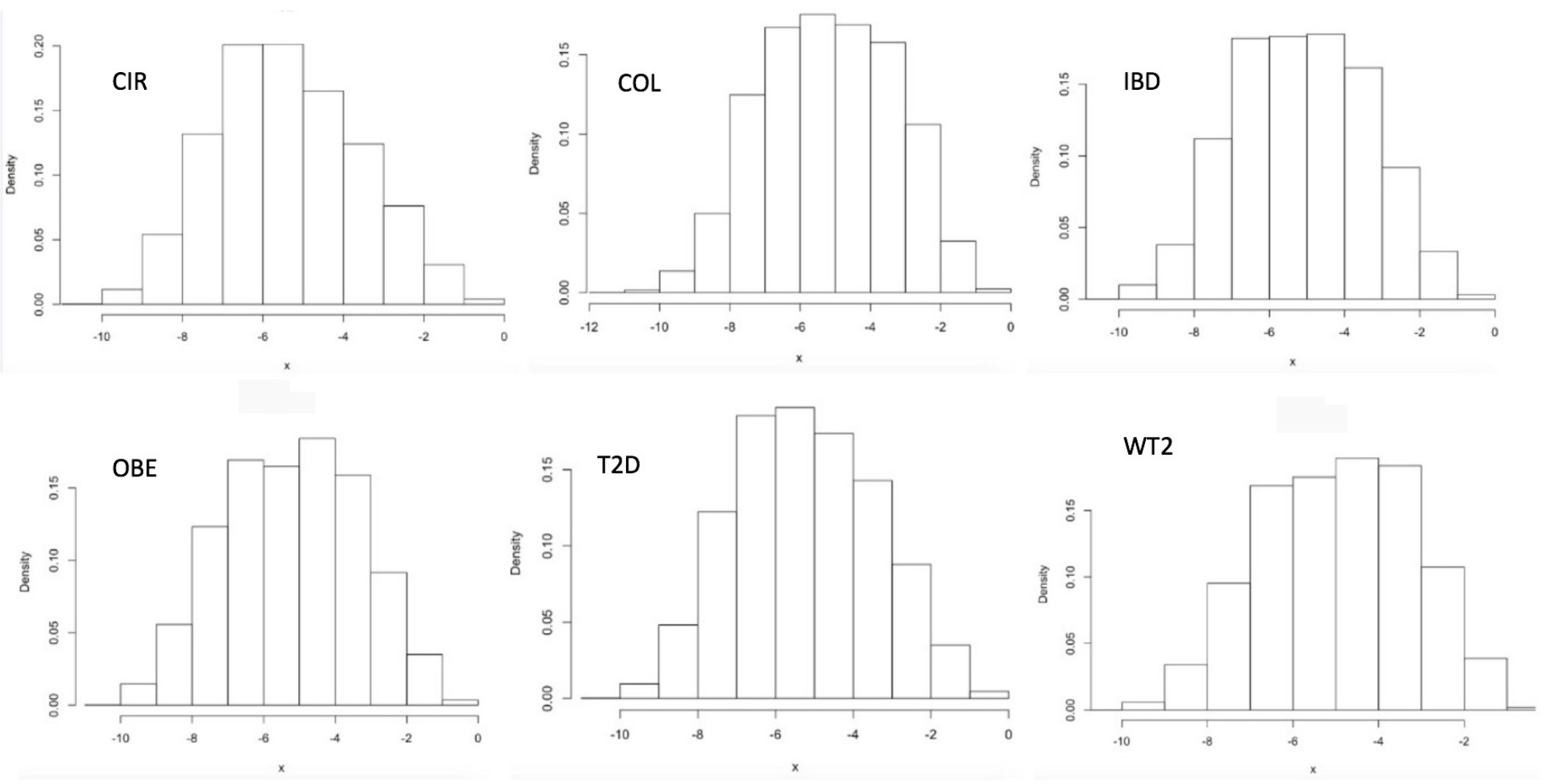}
		\caption{Log histogram of each dataset in group A}
		\label{fig:loghis6}
\end{figure}
In Fig. \ref{fig:hiswhole6}, \textit{left} shows the histogram of the original data. We notice the Zero-Inflated distribution as typically metagenomic data are very sparse. On the \textit{right} we notice the log-transformed distribution of the data (eliminated values of 0) computed by Logarithm (base 4) which is more normally-distributed. In logarithmic scale, the width of each break is 1 being equivalent to a 4-fold increase from the previous bin. As observed from Fig. \ref{fig:loghis6}, histograms of six datasets of group A with Logarithmic scale (base 4) share the same distributions. From our observations, we propose a hypothesis that the models will perform better with such breaks owning values of breaks from $0, 10^{-7}, 4*10^{-7}, 1.6*10^{-6},..., 0.0065536, 1$. The first break is from 0 to $10^{-7}$ which is the minimum value of species abundance known in 6 datasets of group A, each multiplies four times preceding one. We called this binning $``$SPecies Bins$"$  (\textbf{SPB}) in experiments.

\subsubsection{Binning based on Quantile Transformation (QTF)}
We proposed another approach to bin the data, based on a scaling factor which is learned in the training set and then applied to the test set. With different distributions of data, standardization is a commonly-used technique for numerous ML algorithms. Quantile TransFormation (QTF), a Non-Linear transformation, is considered as a strong preprocessing technique because of reducing the outliers effect. Values in new/unseen data (for example, test/validation set) which are lower or higher the fitted range will be set to the bounds of the output distribution. In the experiments, we use this transformation to fit the features' signal to a uniform distribution. These implementations are provided from the scikit-learn library in Python \cite{Garreta2013-ds}.

\subsubsection{Binary Bins}
Beside the previous two methods described above, we also use the binary bins to indicate PResence/absence (PR), respectively corresponding to black/white (BW) values in subsequent images. Note that this method is equivalent to one-hot encoding.
\begin{figure}
\includegraphics[width=0.5\textwidth]{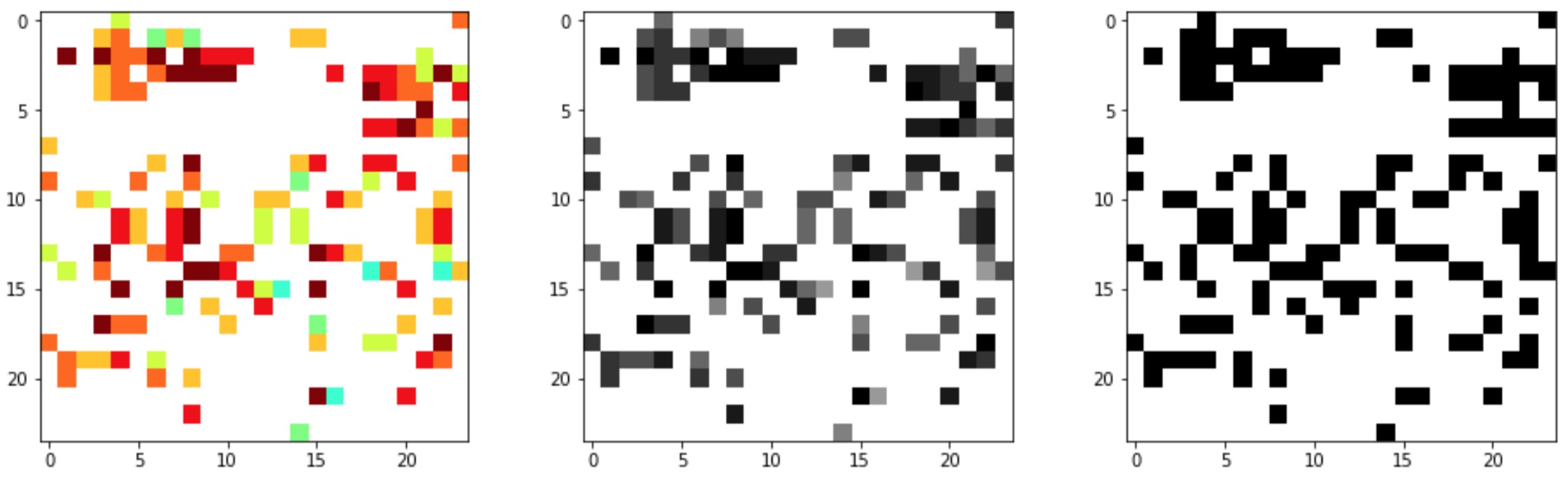}
		\caption{Examples of Fill-up images, Left-Right (types of colors): heatmap, gray scale and black/white}
		\label{fig:fill-up}
\end{figure}

\begin{figure}
	\includegraphics[width=0.5\textwidth]{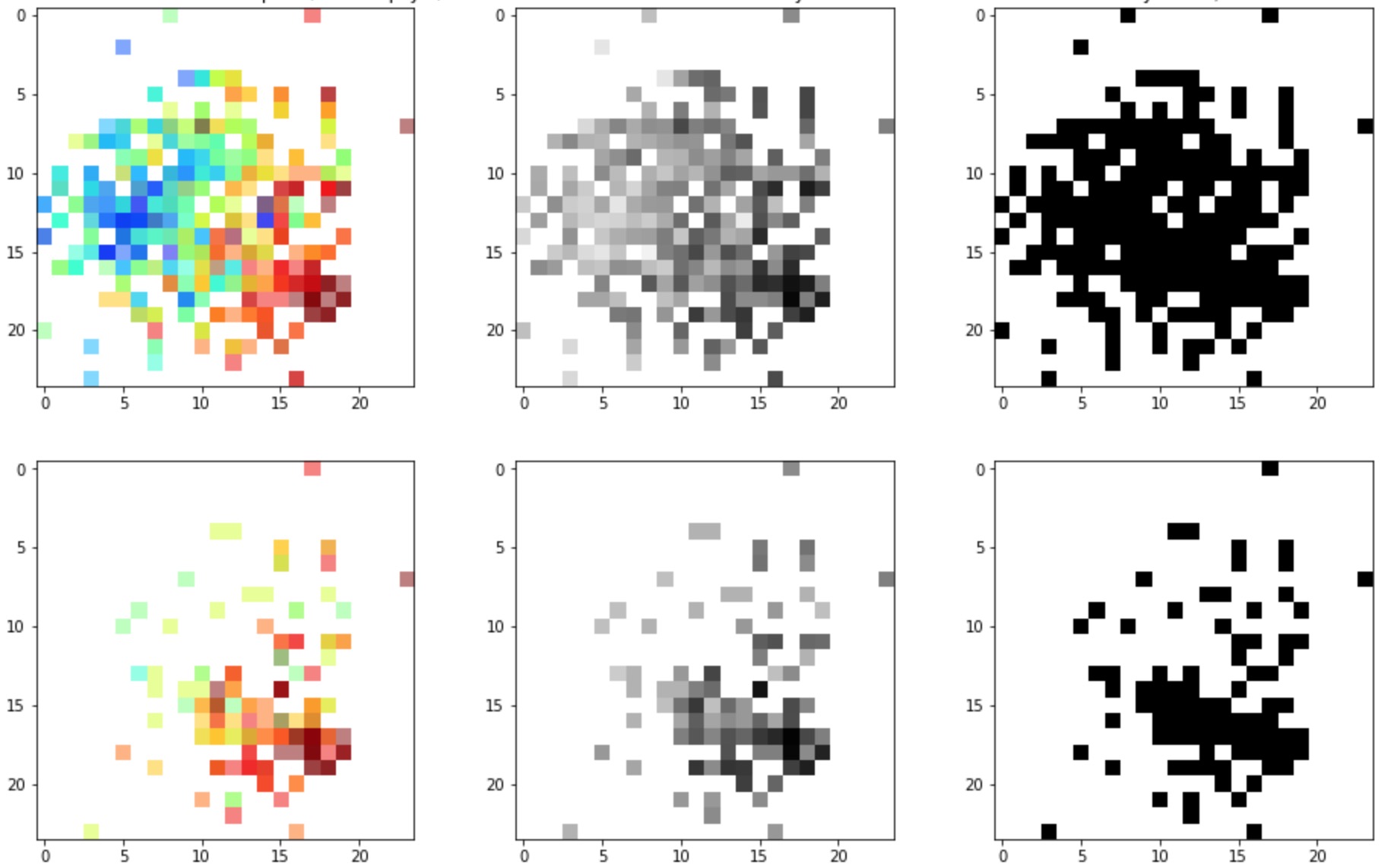}
		\caption{Examples of t-SNE representation, Left-Right (types of images): color images (using heatmap), gray images, black/white images; Top-Down: Global t-SNE maps and images of samples created from the global t-SNE map.}
		\label{fig:tsne}
	\end{figure}

\subsection{Generation of artificial metagenomic images: Fill-up and t-SNE}
\label{method:Image_generation}
\textbf{Fill-up}:\ images are created by arranging abundance/presence values into a matrix in a right-to-left order by row top-to-bottom. The image is square and empty bottom-left of the image are set to zero (white). As an example for a dataset containing 542 features (i.e. bacterial species) in the cirrhosis dataset, we need a matrix of 24$\times$24 to fill up 542 values of species into this square. The first row of pixel is arranged from the first species to the 24th species, the second row includes from the 25th to the 48th and so on till the end. We use distinct colors in binning scale with SPB, QTF and PR to illustrate abundance values of species and black and white for presence/ absence, where white represents absent values (see examples at Fig. \ref{fig:fill-up})
\vspace{\baselineskip}

\textbf{T-SNE maps}: are built based on training sets from raw data with perplexity p=5, learning rate = 200 (default), and the number of iterations set to 300. These maps are used to generate images for training and testing sets. Each species is considered as a point in the map and only species that are present are showed either in abundance or presence using the same color schemes as above (Fig. \ref{fig:tsne}).\par

\subsection{Convolutional Neural Network architectures and models used in the experiments}
\subsubsection*{One-dimensional case}
In order to predict disease using 1D data, we use a Fully Connected neural network (FC) and one 1D convolutional neural network (CNN1D). FC model includes one fully-connected layer and gives one output. This a very simple model but the performance is relatively high. The structure of this network contains a fully connected layer with sigmoid function. CNN1D includes one 1D convolutional layer with 64 filters and one max pooling of 2. We also use classical learning algorithms such RF (50 trees in the forest) and SVM (kernels of Sigmoid, Radial, Linear) for 1D data.

\subsubsection*{Two-dimensional case}
\begin{figure}
\includegraphics[width=0.5\textwidth]{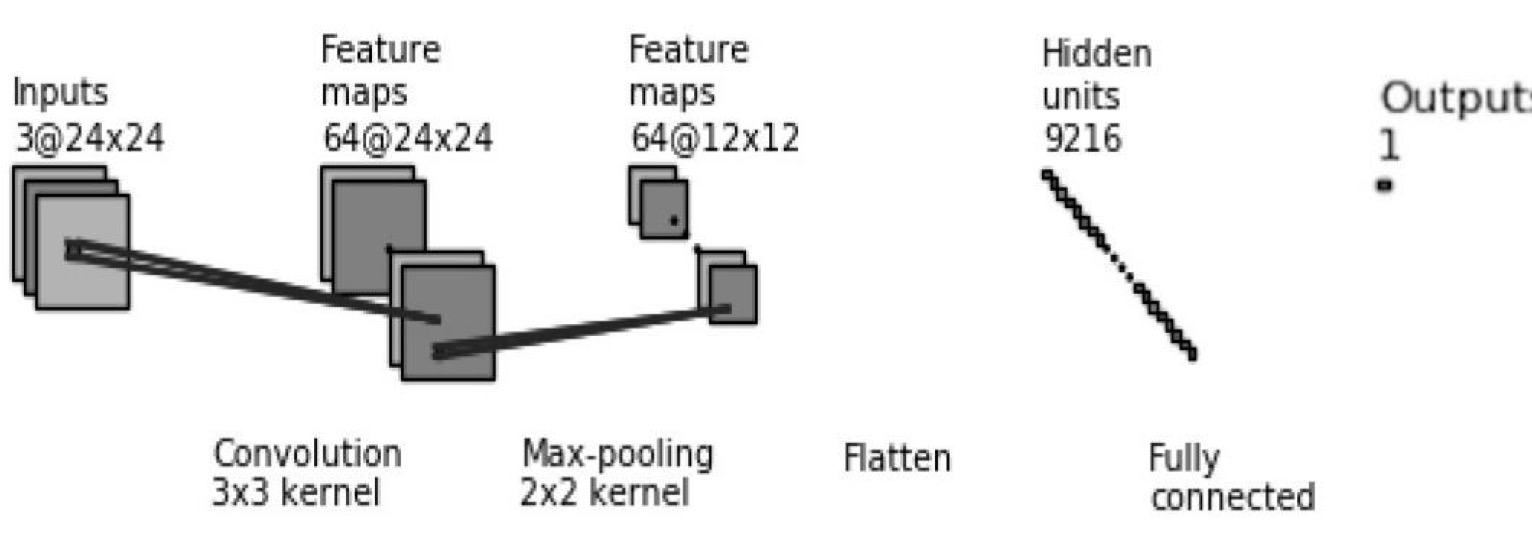}
		\caption{The CNN architecture includes a stack of one convolutional layer with 64 filters of 3x3 and a max pooling of 2x2, followed by one fully connected layer.}
		\label{fig:CNNs_arc}
\end{figure}

\addcontentsline{toc}{subsection}{For 2D data}
Images in Fill-up vary in size from 16$\times$16 to 24$\times$24 depending on the number of features while, in t-SNE, we use 24x24 images for all datasets. The network receives either color images with three channels or gray images with one channel, then passing through a stack of one convolutional layers with 64 kernels of 3$\times$3 (stride 1), followed by a max pooling 2$\times$2 (stride 2). ReLU is used after each convolution. Wide convolution is implemented to preserve the dimension of input after passing through the convolutional layer. There are two approaches for binary classification including using either two output neurons (two-node technique) or one output neuron (one-node technique). We use the latter with a sigmoid activation function at the final layer (see the CNN architecture for images of 24$\times$24 in Fig. \ref{fig:CNNs_arc}). 

\subsubsection*{Experimental Setup}
All networks use the same optimization function (Adam \cite{DBLP:journals/corr/KingmaB14}), learning rate = 0.001, batch size = 16, loss function: binary cross entropy, epoch = 500 (using Early Stopping to avoid over-fitting with the number of epoch patience of 5). \textbf{Met2Img} is implemented in Keras 2.1.3, Tensorflow 1.5 using Python/Scikit-Learn\cite{Garreta2013-ds} and can run either in CPU or GPU architectures. 

\section{Results}
\label{sec:Results}

\subsection{Comparing to the-state-of-the-art (MetAML) \cite{Pasolli2016-dk}}
We run a one-tailed t-test to find significant improvements compared to the-state-of-the-art. The \textit{p-values} are computed based on function tsum.test (\textbf{PASWR}- package for Probability and Statistics with R) to compare results from MetAML. The results which have \textit{p-values} $<$ 0.05 are significant improvements. We use accuracy (ACC) to measure model performances. Classification accuracies were assessed by 10-fold cross validation, repeated and averaged on 10 independent runs using stratified cross validation approach. 

\subsubsection{The results of 1D data}
\begin{table*}[h]
 			\centering
\begin{tabular}{|l|l|l|l|l|l|l|l|l|l|l|l}
\hline
Framework &	Model &	CIR &	COL &	IBD &	OBE &	T2D &	WT2 &	AVG \\ \hline
MetAML &	RF &	0.877 &	0.805 &	0.809 &	0.644 &	0.664 &	0.703 &	0.750 \\ \hline
 &	SVM &	0.834 &	0.743 &	0.809 &	0.636 &	0.613 &	0.596 &	0.705 \\ \hline
Met2Img &	RF &	0.877 &	0.812 &	0.808 &	0.645 &	0.672 &	0.703 &	0.753 \\ \hline
 &	SVM- Sigmoid &	0.509 &	0.603 &	0.775 &	0.648 &	0.515 &	0.553 &	0.600 \\ \hline
 &	SVM- Radial &	0.529 &	0.603 &	0.775 &	0.648 &	0.593 &	0.553 &	0.617 \\ \hline
 &	SVM- Linear &	0.766 &	0.666 &	0.792 &	0.612 &	0.634 &	0.676 &	0.691 \\ \hline
 &	FC &	0.776 &	0.685 &	0.775 &	\textbf{0.656} &	0.665 &	0.607 &	0.694 \\ \hline
 &	CNN1D &	0.775 &	0.722 &	\textbf{0.842} &	\textbf{0.663} &	0.668 &	0.618 &	0.715 \\ \hline
\end{tabular}
  \caption{Performance (in ACC) comparison on 1D data. The significant results are reported in \textbf{bold}. The average performance of six datasets is computed and shown in the last column (AVG)}
  \label{tab:res_1d}
\end{table*}
As shown in the  Table \ref{tab:res_1d}, RF both in MetAML and our framework outperforms other models. In 1D data, CNN1D gives better FC while results for the SVM models are the worst. In SVM models, Linear kernel shows the best performance. From these results, standard and shallow learning algorithms (such as RF) are more robust than deep learning for 1D data.

\subsubsection{The results of 2D data}

\begin{figure}[h]
\centering
\includegraphics[width=0.45\textwidth]{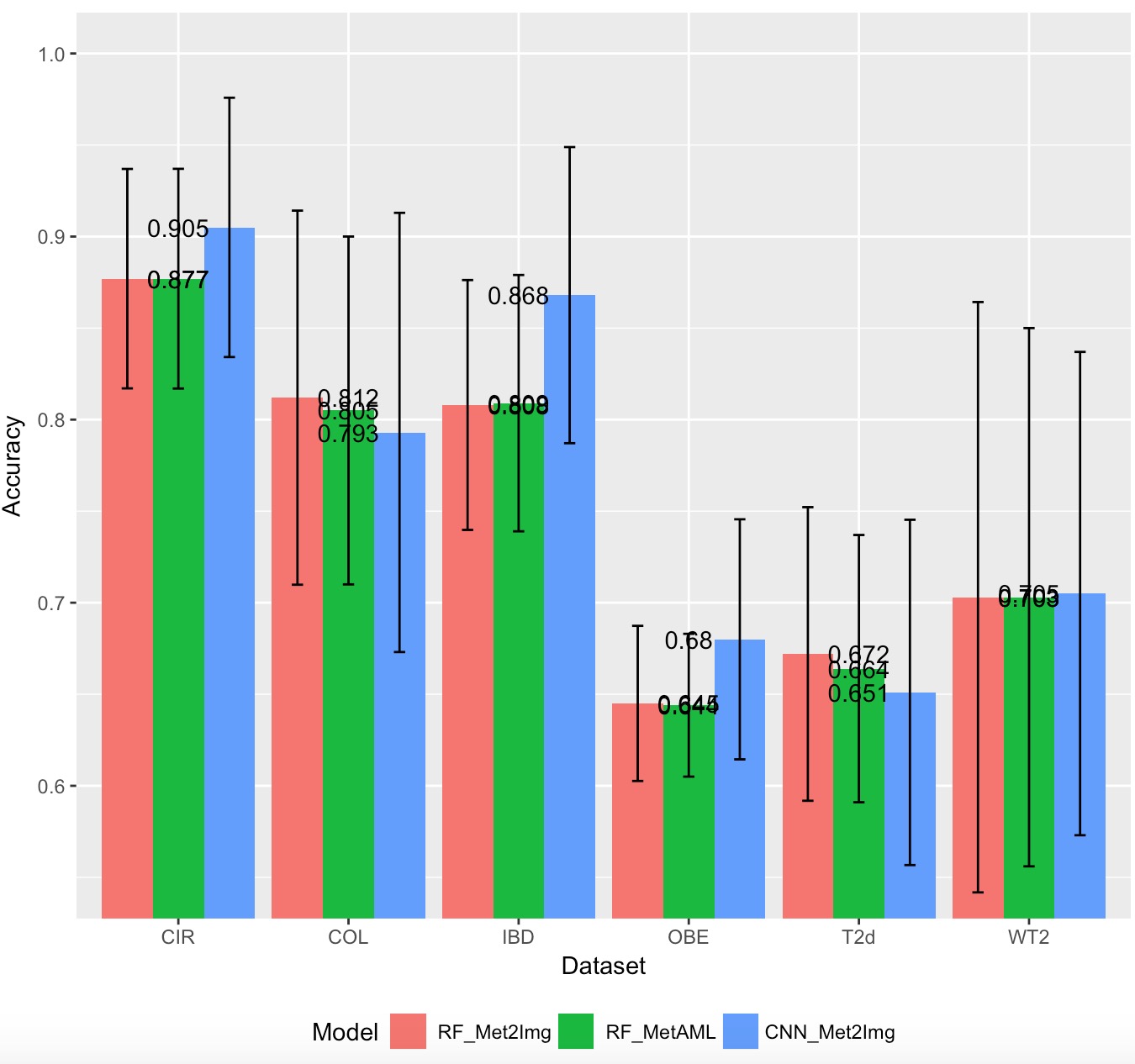}
		\caption{Performance comparison in ACC between CNN model (using Fill-up with SPB and gray images) and shallow learning algorithm (RF) of MetAML and our framework. The standard deviations are shown in error bars.}
		\label{fig:perform_rf_cnn}
	\end{figure}

\begin{table*}
 			\centering
\begin{tabular}{|l|l|l|l|l|l|l|l|l|l|l|l}
 \hline
 &	Bins &	Model &	Color &	\textbf{CIR}  &	\textbf{COL}  &	\textbf{IBD}  &	\textbf{OBE}  &	\textbf{T2D}  &	\textbf{WT2}  &	AVG \\ \hline 
\multicolumn{4}{|c|}{MetAML -   RF} &0.877 &	0.805 &	0.809 &	0.644 &	0.664 &	0.703 &	0.750 \\ \hline
\multicolumn{4}{|c|}{Met2Img -   RF} &0.877 &	0.812 &	0.808 &	0.645 &	0.672 &	0.703 &	0.753 \\ \hline
Fill-up &	PR &	CNN &	BW &	0.880 &	0.763 &	\textbf{0.841} &	\textbf{0.669} &	0.666 &	0.667 &	0.748 \\ \hline
Fill-up &	QTF &	CNN &	color &	\textbf{0.897} &	0.781 &	\textbf{0.837} &	\textbf{0.659} &	0.664 &	0.690 &	0.755 \\ \hline

Fill-up &	SPB &	CNN &	gray &	\textbf{0.905} &	0.793 &	\textbf{0.868} &	\textbf{0.680} &	0.651 &	0.705 &	0.767 \\ \hline
Fill-up &	SPB &	CNN &	color &	\textbf{0.903} &	0.798 &	\textbf{0.863} &	\textbf{0.681} &	0.649 &	0.713 &	0.768 \\ \hline
Fill-up &	PR &	FC &	BW &	0.863 &	0.735 &	\textbf{0.842} &	\textbf{0.672} &	0.656 &	0.712 &	0.747 \\ \hline
Fill-up &	QTF &	FC &	color &	0.887 &	0.765 &	\textbf{0.853} &	\textbf{0.657} &	0.640 &	0.711 &	0.752 \\ \hline
Fill-up &	SPB &	FC &	grays &	0.888 &	0.772 &	\textbf{0.847} &	\textbf{0.686} &	0.652 &	0.716 &	0.760 \\ \hline
Fill-up &	SPB &	FC &	color &	\textbf{0.905} &	0.794 &	\textbf{0.837} &	\textbf{0.679} &	0.659 &	0.713 &	0.764 \\ \hline
tsne &	PR &	CNN &	BW &	0.862 &	0.712 &	0.815 &	\textbf{0.664} &	0.660 &	0.672 &	0.731 \\ \hline
tsne &	QTF &	CNN &	color &	0.878 &	0.746 &	0.811 &	\textbf{0.658} &	\textbf{0.684} &	0.648 &	0.737 \\ \hline
tsne &	QTF &	CNN &	grays &	0.875 &	0.747 &	0.809 &	\textbf{0.664} &	0.672 &	0.651 &	0.736 \\ \hline
tsne &	SPB &	CNN &	grays &	0.877 &	0.761 &	0.809 &	\textbf{0.679} &	0.664 &	0.686 &	0.746 \\ \hline
tsne &	SPB &	CNN &	color &	0.886 &	0.769 &	0.802 &	\textbf{0.682} &	0.658 &	0.685 &	0.747 \\ \hline
tsne &	PR &	FC &	BW &	0.857 &	0.710 &	0.813 &	\textbf{0.662} &	0.640 &	0.688 &	0.728 \\ \hline
tsne &	QTF &	FC &	grays &	0.873 &	0.764 &	0.787 &	\textbf{0.664} &	0.647 &	0.686 &	0.737 \\ \hline
tsne &	QTF &	FC &	color &	0.889 &	0.770 &	\textbf{0.825} &	\textbf{0.660} &	0.638 &	0.650 &	0.739 \\ \hline
tsne &	SPB &	FC &	grays &	0.871 &	0.770 &	0.790 &	\textbf{0.678} &	0.633 &	0.702 &	0.741 \\ \hline
tsne &	SPB &	FC &	color &	\textbf{0.890} &	0.778 &	0.820 &	\textbf{0.689} &	0.615 &	0.705 &	0.749 \\ \hline
\end{tabular}
  \caption{Performance (in ACC) comparison on 2D data. The significant results are reported in \textbf{bold}. The average performance of six datasets is computed and shown in the last column (AVG)}
  \label{tab:res_2d}
\end{table*}
The performances of MetAML is shown in Table \ref{tab:res_2d} are results of RF model, the best model in MetAML. The results with QTF bins share the same pattern with $``$Species bins$"$  (SPB), but the performance of QTF is slightly lower than \textbf{SPB}. Noteworthy, t-SNE using QTF gives significant results for T2D while Fill-up shows poor performance on this dataset. The performance of Fill-up is better compared to t-SNE. In addition, t-SNE performs worse due to many overlapped points, while in contrast every features in Fill-up is visible. As shown in Fig. \ref{fig:perform_rf_cnn}, the CNN model achieves either significant results with CIR, IBD, OBE (\textit{p-value} $<0.0005$) or comparative performance to the others including COL, T2D, WT2 compared to the best model (RF) in MetAML. This proves that CNN is a promising approach for metagenomic data.

\subsection{Applying Met2Img on Sokol’s lab data \cite{Sokol2016-sw}}
In this section, we evaluate the performance of Met2Img on Sokol’s lab data \cite{Sokol2016-sw}, which used to evaluate Ph-CNN in \cite{Fioravanti2017-tr}. The authors performed the six classification tasks on this data to classify HS versus the six partitions CDf, CDr, iCDf, iCDr, UCf, UCr. They proposed Ph-CNN considered as a novel DL architecture using phylogenetic structure of metagenomics for classification tasks. The structure of the model includes a stack of two 1D convolutional layers of 4x16 and a max pooling of 2x1, followed by a fully connected layer with 64 neurons and a dropout of 0.25.
In order to compare to Ph-CNN, we use the same experimental procedure with 5-fold stratified cross validation repeated 10 times (internal validation) and compute Matthews Correlation Coefficient (MCC) as a performance measurement in \cite{Fioravanti2017-tr}, then applying the best model to the external validation datasets. Internal and external validation sets evaluated in our experiments are the same as in \cite{Fioravanti2017-tr}. 

\begin{figure}
\includegraphics[width=0.5\textwidth]{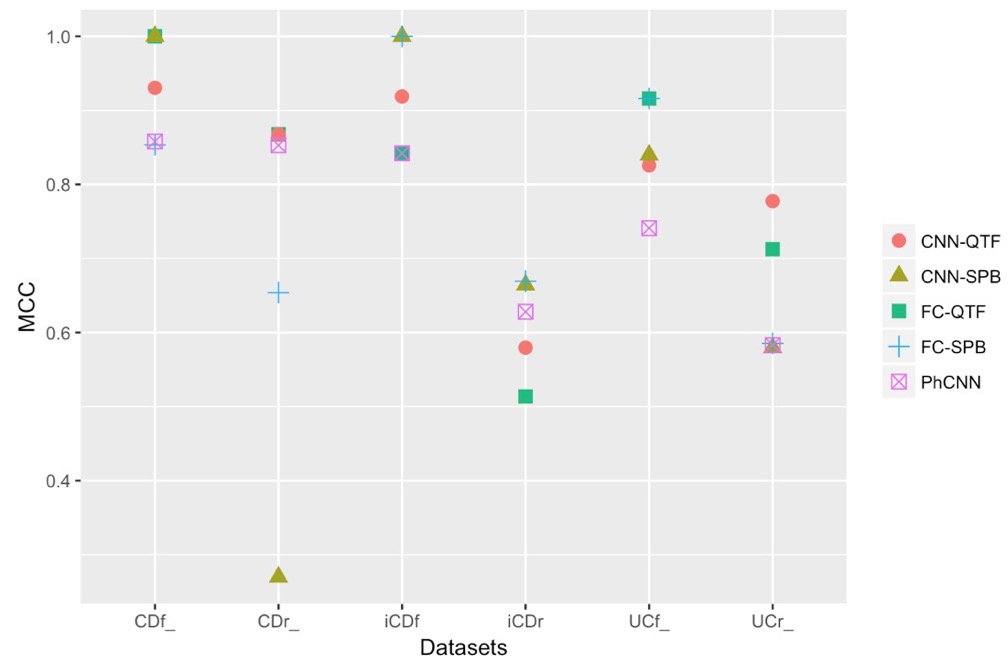}
\caption{Performance comparison in MCC between our approach and Ph-CNN on external validation sets.}
		\label{fig:perform_ph_exte}
\end{figure}

The performance in Tab. \ref{tab:res_ext} and Fig. \ref{fig:perform_ph_exte} using Fill-up with heatmap color images shows significant results compared to Ph-CNN. Especially, Fill-up using QTF outperforms Ph-CNN in internal validation and get better performance in 4 and 5 out of 6 external validation sets. SPB is conducted from species abundance distribution also shows promising results with 4-5 significant results on internal validation sets and achieves greater performances in 4 external validation sets. Noteworthy, for QTF, although CNN model shows worse performance compared to FC, this model reveals encouraging results with better performance on 5/6 external validation sets.

\begin{table*}[h]
 			\centering
\begin{tabular}{|l|l|r|r|r|r|r|r|r|r|r|r|}
\hline

	Model &	Bins &	CDf &	CDr &	iCDf &	iCDr &	UCf& 	UCr& 	AVG \\ \hline
	\multicolumn{2}{|l|}{Internal Validation}	&	 &	 &	 &	 &	& &\\ \hline	
\multicolumn{2}{|c|}{Ph-CNN} &	0.630 &	0.241 &	0.704 &	0.556 &	0.668 &	0.464 &	0.544 \\ \hline
	CNN &	QTF &	0.694 &	\textbf{0.357} &	0.746 &	0.618 &	\textbf{0.811} &	0.467 &	0.616 \\ \hline
		CNN &	SPB &	\textbf{0.808} &	\textbf{0.369} &	\textbf{0.862} &	0.508 &	\textbf{0.790} &	\textbf{0.610} &	0.658 \\ \hline
	FC &	QTF &	\textbf{0.743} &	\textbf{0.362} &	\textbf{0.815} &	\textbf{0.642} &	\textbf{0.796} &	0.519 &	0.646 \\ \hline
	FC &	SPB &	\textbf{0.791} &	0.292 &	\textbf{0.829} &	0.531 &	\textbf{0.795} &	\textbf{0.632} &	0.645 \\ \hline
\multicolumn{2}{|l|}{External Validation}	&	 &	 &	 &	 &	& &\\ \hline	
\multicolumn{2}{|c|}{Ph-CNN} &	0.858 &	0.853 &	0.842 &	0.628 &	0.741 &	0.583 &	0.751 \\ \hline
CNN &	QTF &	\textbf{\textit{0.930}} &	\textbf{\textit{0.868}} &	\textbf{\textit{0.919}} &	0.580 &	\textbf{\textit{0.826}} &	\textbf{\textit{0.777}} &	0.817 \\ \hline
CNN &	SPB &	\textbf{\textit{1.000}} &	0.270 &	\textbf{\textit{1.000}} &	\textbf{\textit{0.664}} &	\textbf{\textit{0.840}} &	0.580 &	0.726 \\ \hline

	FC &	QTF &	\textbf{\textit{1.000}} &	\textbf{\textit{0.868}} &	0.842 &	0.514 &	\textbf{\textit{0.916}} &	\textbf{\textit{0.713}} &	0.809 \\ \hline
	FC &	SPB &	0.854 &	0.654 &	\textbf{\textit{1.000}} &	\textbf{\textit{0.669}} &	\textbf{\textit{0.916}} &	\textbf{\textit{0.585}} &	0.780 \\ \hline

\end{tabular}
  \caption{Classification performance (in MCC) compared to Ph-CNN on six classification tasks on IBD.  The better results on external validation sets are formatted \textbf{\textit{bold-Italic}}. The significant results are reported in \textbf{bold}. The average performances are revealed in the last column (AVG)}
  \label{tab:res_ext}
\end{table*}

\section{Conclusion}
\label{sec:conclusion}
In this work, we introduced a novel approach \textbf{Met2Img} to learn and classify complex metagenomic data  using visual representations constructed with Fill-up and t-SNE. We explore three methods of binning, namely \textbf{SPB}, \textbf{QTF} and \textbf{PR} with various colormaps using heatmap, gray, and black/white images. The bins and images approach are constructed and produced using training data only, thus avoiding over-fitting issues. 

Our results illustrate that the Fill-up approach outperforms the t-SNE. This may be due to several factors. First, the features in the Fill-up are all visible while features in t-SNE are often overlapping. Second, the Fill-up approach integrates prior knowledge on the phylogenetic classification of the features. In addition, the t-SNE images are more complex than the Fill-up images. Noteworthy, with the Fill-up we show significant improvements three data sets, while the t-SNE reveals significant improvement on one data set, the T2D. The FC model outperforms CNN model in color images while CNN model achieves better performance than FC for gray and black/white images. Besides, the representations based on 2D images yield better results compared to 1D data. In general, the proposed Met2Img method outperforms the state-of-the-art both on species and genus abundance data. 

Currently we are investigating various deep learning architectures, and also explore integration of other heterogeneous omics data.
%
%
\bibliography{cap2018}

\end{document}